%% file: main.tex
\documentclass[11pt,a4paper]{article}
\usepackage[hyperref]{acl2019}
\usepackage{times}
\usepackage{latexsym}
\usepackage{url}
\usepackage{amsmath,amssymb}
\usepackage{graphicx}
\graphicspath{{figures/}}
\usepackage{booktabs}
\usepackage{multirow}
\usepackage[normalem]{ulem}
\def\figref#1{Fig.~\ref{#1}}
\def\secref#1{Sec.~\ref{#1}}
\def\tabref#1{Table~\ref{#1}}
\def\eqnref#1{Eqn.~\ref{#1}}

\aclfinalcopy 

\title{Avoiding Reasoning Shortcuts: Adversarial Evaluation, Training, and Model Development for Multi-Hop QA}

\author{Yichen Jiang \and Mohit Bansal \\
  UNC Chapel Hill \\
  {\tt \{yichenj, mbansal\}@cs.unc.edu} \\
 }

\date{}

\begin{document}
\maketitle

\input{tex/abstract}
\input{tex/intro}
\input{tex/methods}
\input{tex/models}
\input{tex/experiments}
\input{tex/results}

\input{tex/analysis}
\input{tex/related}
\input{tex/conclusion}
\input{tex/acknowledgement}

\bibliography{main}
\bibliographystyle{acl_natbib}

\input{tex/appendix}

\end{document}

%% file: tex/abstract.tex
\begin{abstract}

Multi-hop question answering requires a model to connect multiple pieces of evidence scattered in a long context to answer the question. In this paper, we show that in the multi-hop HotpotQA~\cite{yang2018hotpotqa} dataset, the examples often contain reasoning shortcuts through which models can directly locate the answer by word-matching the question with a sentence in the context. We demonstrate this issue by constructing adversarial documents that create contradicting answers to the shortcut but do not affect the validity of the original answer. The performance of strong baseline models drops significantly on our adversarial evaluation, indicating that they are indeed exploiting the shortcuts rather than performing multi-hop reasoning. After adversarial training, the baseline's performance improves but is still limited on the adversarial evaluation. Hence, we use a control unit that dynamically attends to the question at different reasoning hops to guide the model's multi-hop reasoning. We show that this 2-hop model trained on the regular data is more robust to the adversaries than the baseline model. After adversarial training, this 2-hop model not only achieves improvements over its counterpart trained on regular data, but also outperforms the adversarially-trained 1-hop baseline. We hope that these insights and initial improvements will motivate the development of new models that combine explicit compositional reasoning with adversarial training.\footnote{Our code and data are publicly available at: \\ \url{https://github.com/jiangycTarheel/Adversarial-MultiHopQA}}
\end{abstract}

%% file: tex/intro.tex
\section{Introduction}
\begin{figure}[t]
\centering
\includegraphics[width=0.47\textwidth]{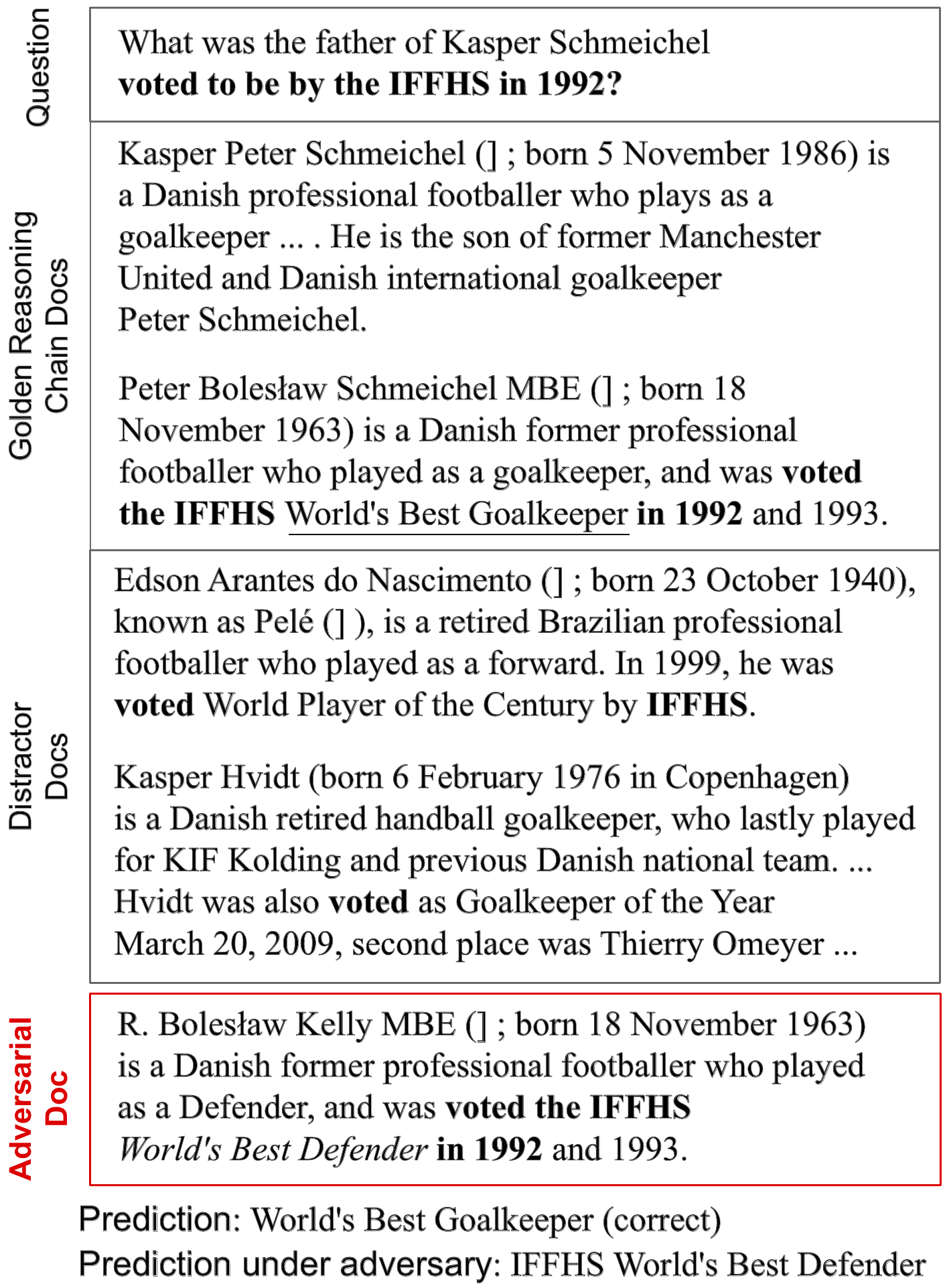}
\vspace{-1pt} 
\caption{HotpotQA example with a reasoning shortcut, and our adversarial document that eliminates this shortcut to necessitate multi-hop reasoning. 
 \vspace{-1pt}
\label{fig:example-intro}}
\end{figure}

The task of question answering (QA) requires the model to answer a natural language question by finding relevant information in a given natural language context.
Most QA datasets require single-hop reasoning only, which means that the evidence necessary to answer the question is concentrated in a single sentence or located closely in a single paragraph (Q: ``What's the color of the sky?'', Context: ``The sky is blue.'', Answer: ``Blue'').
Such datasets emphasize the role of matching and aligning information between the question and the context (``sky$\rightarrow$sky, color$\rightarrow$blue''). Previous works have shown that models with strong question-aware context representation \citep{seo2016bidaf,xiong2016dynamic} can achieve super-human performance on single-hop QA tasks like SQuAD \citep{rajpurkar2016squad,rajpurkar2018squad2}.

Recently, several multi-hop QA datasets, such as QAngaroo~\citep{welbl2017qangaroo} and HotpotQA~\citep{yang2018hotpotqa}, have been proposed to further assess QA systems' ability to perform composite reasoning. In this setting, the information required to answer the question is scattered in the long context and the model has to connect multiple evidence pieces to pinpoint to the final answer. \figref{fig:example-intro} shows an example from the HotpotQA dev set, where it is necessary to consider information in two documents to infer the hidden reasoning chain ``Kasper Schemeichel $\xrightarrow{son\_of}$ Peter Schemeichel $\xrightarrow{voted\_as}$ World's Best Goalkeeper" that leads to the final answer.
However, in this example, one may also arrive at the correct answer by matching a few keywords in the question (``voted, IFFHS, in 1992") with the corresponding fact in the context without reasoning through the first hop to find ``father of Kasper Schmeichel", as neither of the two distractor documents contains sufficient distracting information about another person ``voted as something by IFFHS in 1992".
Therefore, a model performing well on the existing evaluation does not necessarily suggest its strong compositional reasoning ability.
To truly promote and evaluate a model's ability to perform multi-hop reasoning, there should be no such ``reasoning shortcut'' where the model can locate the answer with single-hop reasoning only. 
This is a common pitfall when collecting multi-hop examples and is difficult to address properly.

In this work, we improve the original HotpotQA distractor setting\footnote{HotpotQA has a fullwiki setting as an open-domain QA task. In this work, we focus on the distractor setting as it provides a less noisy environment to study machine reasoning.} by adversarially generating better distractor documents that make it necessary to perform multi-hop reasoning in order to find the correct answer. 
As shown in \figref{fig:example-intro}, we apply phrase-level perturbations to the answer span and the titles in the supporting documents to create the adversary with a new title and a fake answer to confuse the model. 
With the adversary added to the context, it is no longer possible to locate the correct answer with the single-hop shortcut, which now leads to two possible answers (``World's Best Goalkeeper" and ``World's Best Defender").
We evaluate the strong ``Bi-attention + Self-attention" model~\citep{seo2016bidaf,wang2017gated} from \citet{yang2018hotpotqa} on our constructed adversarial dev set (adv-dev), and find that its EM score drops significantly. In the example in \figref{fig:example-intro}, the model is confused by our adversary and predicts the wrong answer (``World's Best Defender").
Our experiments further reveal that when strong supervision of the supporting facts that contain the evidence is applied, the baseline achieves a significantly higher score on the adversarial dev set.
This is because the strong supervision encourages the model to not only locate the answer but also find the evidence that completes the first reasoning hop and hence promotes robust multi-hop reasoning behavior from the model.
We then train the baseline with supporting fact supervision on our generated adversarial training set (adv-train) and observe significant improvement on adv-dev.
However, the result is still poor compared to the model's performance on the regular dev set because this single-hop model is not well-designed to perform multi-hop reasoning.

To motivate and analyze some new multi-hop reasoning models, we propose an initial architecture by incorporating the recurrent control unit from \citet{hudson2018compositional}, which dynamically computes a distribution over question words at each reasoning hop to guide the multi-hop bi-attention. 
In this way, the model can learn to put the focus on ``father of Kasper Schmeichel" at the first step and then attend to ``voted by IFFHS in 1992" in the second step to complete this 2-hop reasoning chain.
When trained on the regular data, this 2-hop model outperforms the single-hop baseline in the adversarial evaluation, indicating improved robustness against adversaries.
Furthermore, this 2-hop model, with or without supporting-fact supervision, can benefit from adversarial training and achieve better performance on adv-dev compared to the counterpart trained with the regular training set, while also outperforming the adversarially-trained baseline.
Overall, we hope that these insights and initial improvements will motivate the development of new models that combine explicit compositional reasoning with adversarial training.

%% file: tex/methods.tex
\section{Adversarial Evaluation}
\begin{figure*}[t]
\centering
\includegraphics[width=0.98\textwidth]{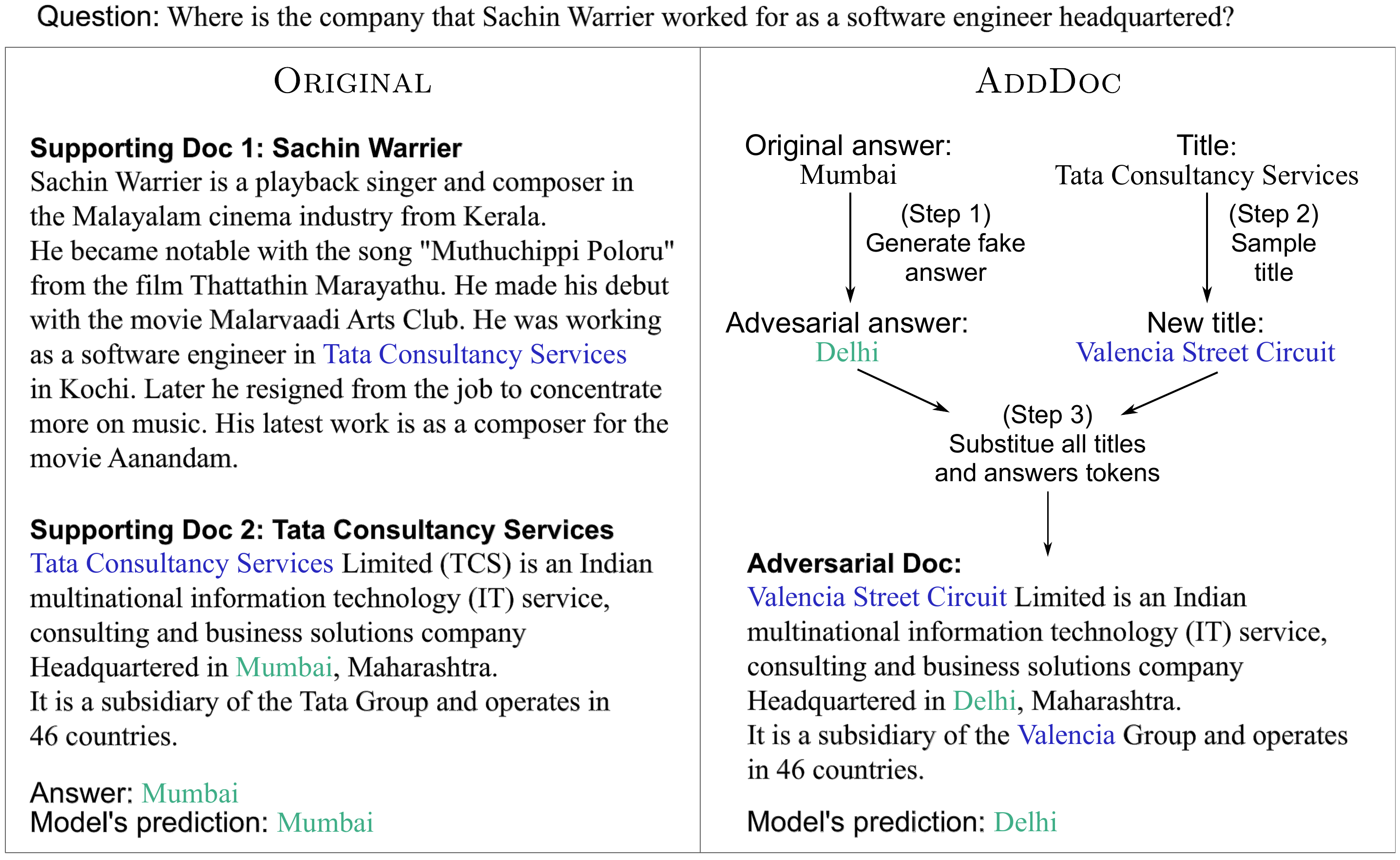}
\vspace{-8pt}
\caption{An illustration of our \textsc{AddDoc} procedure. In this example, the keyword ``headquarter" appears in no distractor documents. Thus the reader can easily infer the answer by looking for this keyword in the context. \vspace{-1pt}
\label{fig:adv-gen}
}
\end{figure*}

\subsection{The HotpotQA Task}
The HotpotQA dataset \citep{yang2018hotpotqa} is composed of 113k human-crafted questions, each of which can be answered with facts from two Wikipedia articles. 
During the construction of the dataset, the crowd workers are asked to come up with questions requiring reasoning about two given documents. 
\citet{yang2018hotpotqa} then select the top-8 documents from Wikipedia with the shortest bigram TF-IDF~\cite{chen2017reading} distance to the question as the distractors to form the context with a total of 10 documents. Since the crowd workers are not provided with distractor documents when generating the question, there is no guarantee that both supporting documents are necessary to infer the answer given the entire context. The multi-hop assumption can be broken by incompetent distractor documents in two ways. 
First, one of the selected distractors may contain all required evidence to infer the answer (e.g., ``The father of Kasper Schmeichel was voted the IFFHS World's Best Goalkeeper in 1992."). Empirically, we find no such cases in HotpotQA, as Wiki article about one subject rarely discusses details of another subject.
Second, the entire pool of distractor documents may not contain the information to truly distract the reader/model.
As shown in \figref{fig:example-intro}, one can directly locate the answer by looking for a few keywords in the question (``voted, IFFHS, in 1992") without actually discovering the intended 2-hop reasoning path.
We call this pattern of bypassing the first reasoning hop the ``reasoning shortcut", and we find such shortcuts exist frequently in the non-comparison-type examples in HotpotQA.\footnote{HotpotQA also includes a subset of comparison questions (e.g.,``Are Leo and Kate of the same age?") that make up to 21\% of total examples in the dev set. These questions can't be answered without aggregating information from multiple documents, as shortcuts like ``Leo is one-year older than Kate" rarely exist in Wikipedia articles. 
Therefore, we simply leave these examples unchanged in our adversarial data.}
We randomly sample 50 ``bridge-type" questions in the dev set, and found that 26 of them contain this kind of reasoning shortcut.

\subsection{Adversary Construction}
\label{ssec:adv-const}
To investigate whether neural models exploit reasoning shortcuts instead of exploring the desired reasoning path, we adapt the original examples in HotpotQA to eliminate these shortcuts.
Given a context-question-answer tuple $(C, q, a)$ that may contain a reasoning shortcut, the objective is to produce $(C', q, a)$ such that (1) $a$ is still the valid answer to the new tuple, (2) $C'$ is close to the original example, and (3) there is no reasoning shortcut that leads to a single answer. In HotpotQA, there is a subset of 2 supporting documents $P \subset C$ that contains all evidence needed to infer the answer. 
To achieve this, we propose an adversary \textsc{AddDoc} (illustrated in \figref{fig:adv-gen}) that constructs documents $P'$ to get $(\xi(C, P'), q, a)$ where $\xi$ is a function that mixes the context and adversaries. 

Suppose $p_2 \in P$ is a document containing the answer $a$ and $p_1 \in P$ is the other supporting document.\footnote{$|P| = 2$ in HotpotQA. If both documents in $P$ contain the answer, we apply \textsc{AddDoc} twice while alternating the choice of $p_1$ and $p_2$} 
\textsc{AddDoc} applies a word/phrase-level perturbation to $p_2$ so that the generated $p'_2$ contains a fake answer that satisfies the reasoning shortcut but does not contradict the answer to the entire question (e.g., the adversarial document in \figref{fig:adv-gen}).
First, for every non-stopword in the answer, we find the substitute within the top-10 closest words in GloVe \citep{pennington2014glove} 100-d vector space that doesn't have an overlapping substring longer than 3 with the original answer (``Mumbai $\rightarrow$ Delhi, Goalkeeper $\rightarrow$ Defender").
If this procedure fails, we randomly sample a candidate from the entire pool of answers in the HotpotQA dev set (e.g., ``Rome" for \figref{fig:adv-gen} or ``defence of the Cathedral" for \figref{fig:example-intro}). 
We then replace the original answer in $p_2$ with our generated answer to get $p'_2$. 
If the original answer spans multiple words, we substitute one non-stopword in the answer with the corresponding sampled answer word to create the fake answer (``World's Best Goalkeeper $\rightarrow$ World's Best Defender") and replace all mentions of the original answer in $p'_2$.

The resulting paragraph $p'_2$ provides an answer that satisfies the reasoning shortcut, but also contradicts the real answer to the entire question as it forms another valid reasoning chain connecting the question to the fake answer (``Sachin Warrier $\xrightarrow{workAt}$ TCS $\xrightarrow{at}$ Delhi").
To break this contradicting reasoning chain, we need to replace the bridge entity that connects the two pieces of evidence (``Tata Consultancy Services" in this case) with another entity so that the generated answer no longer serves as a valid answer to the question.
We replace the title of $p'_2$ with a candidate randomly sampled from all document titles in the HotpotQA dev set.
If the title of $p_1$ appears in $p'_2$, we also replace it with another sampled title to entirely eliminate the connection between $p'_2$ and $p_1$. 
Empirically, we find that the title of either $p_1$ or $p_2$ serves as the bridge entity in most examples.
Note that it is possible that models trained on our adversarial data could simply learn new reasoning shortcuts in these adversaries by ignoring adversarial documents with randomly-sampled titles, because these titles never appear in any other document in the context.
Hence, to eliminate this bias in title occurrence, for each adversarial document, we additionally find another document from the entire dev set that contains the exact title of our adversarial document and add it to the context.\footnote{Empirically, we find that our models trained on the adversarial data without this final title-balancing step do not seem to be exploiting this new shortcut, because they still perform equally well on the title-balanced adversarial evaluation. However, we keep this final title-balancing step in our adversary-generation procedure so as to prevent future model families from exploiting this title shortcut.}
Every new document added to the context replaces an original non-supporting document so that the total number of documents in context remains unchanged. 
Note that \textsc{AddDoc} adversaries are model-independent, which means that they require no access to the model or any training data, similar to the \textsc{AddOneSent} in~\citet{jia2017adversarial}.

%% file: tex/models.tex
\section{Models}

\begin{figure*}[t]
\centering
\includegraphics[width=0.99\textwidth]{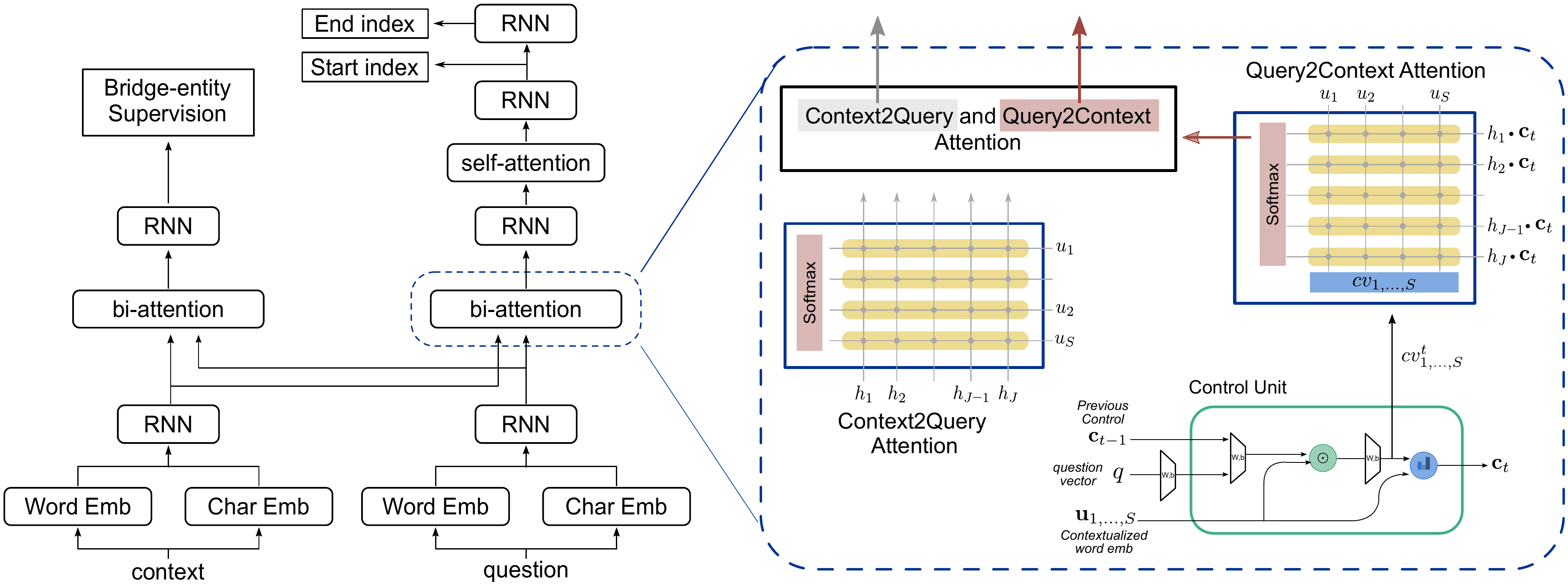}
\vspace{1pt} 
\caption{A 2-hop bi-attention model with a control unit. The Context2Query attention is modeled as in~\citet{seo2016bidaf}. The output distribution $cv$ of the control unit is used to bias the Query2Context attention.
\label{fig:model} 
\vspace{-1pt}}
\end{figure*}

\subsection{Encoding}
We first describe the pre-processing and encoding steps.
We use a Highway Network~\citep{srivastava2015highway} of dimension $v$, which merges the character embedding and GloVe word embedding~\citep{pennington2014glove}, to get the word representations for the context and the question as \(\mathbf{x} \in \mathbb{R}^{J \times v}\) and \(\mathbf{q} \in \mathbb{R}^{S \times v}\) where \(J\) and \(S\) are the lengths of the context and question.
We then apply a bi-directional LSTM-RNN~\citep{hochreiter1997lstm} of $d$ hidden units to get the contextualized word representations for the context and question: $\mathbf{h} = \mathrm{BiLSTM} (\mathbf{x}); \; \mathbf{u} = \mathrm{BiLSTM} (\mathbf{q})$ so that \(\mathbf{h} \in \mathbb{R}^{J \times 2d}\) and \(\mathbf{u} \in \mathbb{R}^{S \times 2d}\). 

\subsection{Single-Hop Baseline}
We use the bi-attention + self-attention model~\cite{yang2018hotpotqa,clark2017simple}, which is a strong  near-state-of-the-art\footnote{At the time of submission: March 3rd, 2019.} model on HotpotQA. 
Given the contextualized encoding $\mathbf{h}, \mathbf{u}$ for the context and question, $\mathrm{BiAttn}(\mathbf{h},\mathbf{u})$~\cite{seo2016bidaf,xiong2016dynamic} first computes a similarity matrix $M^{S\times J}$ between every question and context word and use it to derive context-to-query attention:
\vspace{-5pt}
\begin{equation} \label{eq:c-to-q}
\begin{split}
M_{s,j} &= W_1\mathbf{u}_s + W_2\mathbf{h}_j + W_3(\mathbf{u}_s \odot \mathbf{h}_j)\\
p_{s,j} &= \frac{\mathrm{exp}(M_{s,j})}{\sum_{s=1}^{S}\mathrm{exp}(M_{s,j})}\\
\mathbf{c_q}_j &= \sum_{s=1}^{S} p_{s,j}\mathbf{u}_s
\end{split}
\end{equation}
where $W_1, W_2$ and $W_3$ are trainable parameters, and $\odot$ is element-wise multiplication. Then the query-to-context attention vector is derived as:
\vspace{-5pt}
\begin{equation} \label{eq:q-to-c}
\begin{split}
m_j &= \mathrm{max}_{1\leq s\leq S} \;M_{s,j}\\
p_{j} &= \frac{\mathrm{exp}(m_{j})}{\sum_{j=1}^{J}\mathrm{exp}(m_{j})}\\
\mathbf{q_c} &= \sum_{j=1}^{J} p_{j}\mathbf{h}_j
\end{split}
\end{equation}
We then obtain the question-aware context representation and pass it through another layer of $\mathrm{BiLSTM}$:
\begin{equation} \label{eq:biattn}
\begin{split}
    \mathbf{h'}_j &= [\mathbf{h}_j; \mathbf{c_q}_j; \mathbf{h}_j \odot \mathbf{c_q}_j; \mathbf{c_q}_j \odot \mathbf{q_c}]\\
    \mathbf{h}^1 &= \mathrm{BiLSTM}(\mathbf{h'})
\end{split}
\end{equation}
where ; is concatenation.
Self-attention is modeled upon $\mathbf{h}^1$ as $\mathrm{BiAttn}(\mathbf{h}^1,\mathbf{h}^1)$ to produce $\mathbf{h}^2$.
Then, we apply linear projection to $\mathbf{h}^2$ to get the start index logits for span prediction and the end index logits is modeled as $\mathbf{h}^3 = \mathrm{BiLSTM}(\mathbf{h}^2)$ followed by linear projection.
Furthermore, the model uses a 3-way classifier on $\mathbf{h}^3$ to predict the answer as ``yes", ``no", or a text span. 
The model is additionally supervised to predict the sentence-level supporting fact by applying a binary classifier to every sentence on $\mathbf{h}^2$ after self-attention.

\subsection{Compositional Attention over Question}
To present some initial model insights for future community research, we try to improve the model's ability to perform composite reasoning using a recurrent control unit \cite{hudson2018compositional} that computes a distribution-over-word on the question at each hop.
Intuitively, the control unit imitates human's behavior when answering a question that requires multiple reasoning steps. 
For the example in \figref{fig:example-intro}, a human reader would first look for the name of ``Kasper Schmeichel's father". Then s/he can locate the correct answer by finding what ``Peter Schmeichel" (the answer to the first reasoning hop) was ``voted to be by the IFFHS in 1992".
Recall that $S,J$ are the lengths of the question and context. At each hop $i$, given the recurrent control state $c_{i-1}$, contextualized question representation $\mathbf{u}$, and question's vector representation $q$, the control unit outputs a distribution $cv$ over all words in the question and updates the state $c_{i}$:
\begin{equation}\label{eq:control}
\begin{split}
cq_i & = \mathrm{Proj}[c_{i-1};q]; \;\;\; ca_{i,s} = \mathrm{Proj}(cq_i \odot \mathbf{u}_s)\\
cv_{is} & = \mathrm{softmax}(ca_{is}); \;\;\; c_i = \sum_{s=1}^S cv_{i,s} \cdot \mathbf{u}_s
\end{split}
\end{equation}
where $\mathrm{Proj}$ is the linear projection layer. 
The distribution $cv$ tells which part of the question is related to the current reasoning hop.

Then we use $cv$ and $c_i$ to bias the $\mathrm{BiAttn}$ described in the single-hop baseline. Specifically, we use $\mathbf{h}\odot c_i$ to replace $\mathbf{h}$ in \eqnref{eq:c-to-q}, \eqnref{eq:q-to-c}, and \eqnref{eq:biattn}.
Moreover, after we compute the similarity matrix $M$ between question and context words as in \eqnref{eq:c-to-q}, instead of max-pooling $M$ on the question dimension (as done in the single-hop bi-attention), we calculate the distribution over $J$ context words as:
\vspace{-5pt}
\begin{equation} \label{eq:weighted-biattn}
\begin{split}
m'_j &= cv \cdot M\\
p_{j} &= \frac{\mathrm{exp}(m'_{j})}{\sum_{j=1}^{J}\mathrm{exp}(m'_{j})}\\
\mathbf{q_c} &= \sum_{j=1}^{J} p_{j}\mathbf{h}_j
\end{split}
\end{equation}

The query-to-context attention vector $\mathbf{q_c}$ is applied to the following computation in \eqnref{eq:biattn} to get the query-aware context representation.
Here, with the output distribution from the control unit, $\mathbf{q_c}$ represents the context information that is most relevant to the sub-question of the current reasoning hop, as opposed to encoding the context most related to any question word in the original bi-attention.
Overall, this model (illustrated in \figref{fig:model}) combines the control unit from the state-of-the-art multi-hop VQA model and the widely-adopted bi-attention mechanism from text-based QA to perform composite reasoning on the context and question.

\paragraph{Bridge Entity Supervision}
However, even with the multi-hop architecture to capture a hop-specific distribution over the question, there is no supervision on the control unit's output distribution $cv$ about which part of the question is important to the current reasoning step, thus preventing the control unit from learning the composite reasoning skill.
To address this problem, we look for the bridge entity (defined in \secref{ssec:adv-const}) that connects the two supporting documents.
We supervise the main model to predict the bridge entity span (``Tata Consultancy Services" in \figref{fig:adv-gen}) after the first bi-attention layer, which indirectly encourages the control unit to look for question information related to this entity (``company that Sachin Warrier worked for as a software engineer") at the first hop.
For examples with the answer appearing in both supporting documents,\footnote{This mostly happens for questions requiring checking multiple facts of an entity.} the intermediate supervision is given as the answer appearing in the first supporting document, while the answer in the second supporting document serves as the answer-prediction supervision.

%% file: tex/experiments.tex
\section{Experimental Setup}
\label{sec:exp}
\paragraph{Adversarial Evaluation and Training}
\label{ssec:exp-adv-eval-train}
For all the adversarial analysis in this paper, we construct four adversarial dev sets with different numbers of adversarial documents per supporting document containing answer (4 or 8) and mixing strategy (randomly insert or prepend). We name these 4 dev sets ``Add4Docs-Rand", ``Add4Docs-Prep", ``Add8Docs-Rand", and ``Add8Docs-Prep".
For adversarial training, we choose the ``Add4Docs-Rand" training set since it is shown in \citet{wang2018robust} that training with randomly inserted adversaries yields the model that is the most robust to the various adversarial evaluation settings. 
In the adversarial training examples, the fake titles and answers are sampled from the original training set.
We randomly select 40\% of the adversarial examples and add them to the regular training set to build our adversarial training set.

\paragraph{Dataset and Metrics}
We use the HotpotQA~\cite{yang2018hotpotqa} dataset's distractor setting.
We show EM scores rather than F1 scores because our generated fake answer usually has word-overlap with the original answer, but the overall result trends and take-away's are the same even for F1 scores. 

\paragraph{Training Details}
We use 300-d pre-trained GloVe word embedding~\cite{pennington2014glove} and 80-d encoding LSTM-RNNs.
The control unit of the 2-hop model has an 128-d internal state.
We train the models using Adam~\cite{Kingma2014AdamAM} optimizer, with an initial learning rate of 0.001.
We keep exponential moving averages of all trainable variables in our models and use them during the evaluation.

%% file: tex/results.tex
\section{Results}
\begin{table}[t]
\centering
\begin{small}
\begin{tabular}[t]{lcccc}
\toprule
Train & Reg & Reg & Adv & Adv\\
Eval & Reg & Adv & Reg & Adv \\
\midrule
    1-hop Base & 42.32 & 26.67 & 41.55 & 37.65 \\
    1-hop Base + sp & 43.12 & 34.00 & 45.12 & 44.65 \\
    2-hop & \textbf{47.68} & \textbf{34.71} & 45.71 & 40.72 \\
    2-hop + sp & 46.41 & 32.30 & \textbf{47.08} & \textbf{46.87} \\
\bottomrule
\end{tabular}
\caption{EM scores after training on the regular data or on the adversarial training set \textsc{Add4Docs-Rand}, and evaluation on the regular dev set or the \textsc{Add4Docs-Rand} adv-dev set. ``1-hop Base" and "2-hop" do not have sentence-level supporting-facts supervision.
    \vspace{-5pt}\label{tab:main}}
\end{small}
\end{table}

\paragraph{Regularly-Trained Models}
In our main experiment, we compare four models' performance on the regular HotpotQA  and Add4Docs-Rand dev sets, when trained on two different training sets (regular or adversarial), respectively.
The first two columns in \tabref{tab:main} show the result of models trained on the regular training set only. 
As shown in the first row, the single-hop baseline trained on regular data performs poorly on the adversarial evaluation, suggesting that it is indeed exploiting the reasoning shortcuts instead of actually performing the multi-hop reasoning in locating the answer.
After we add the supporting fact supervision (2nd row in \tabref{tab:main}), we observe a significant improvement\footnote{All stat. signif. is based on bootstrapped randomization test with 100K samples~\citep{Efron:94}.} (p $< 0.001$) on the adversarial evaluation, compared to the baseline without this strong supervision.
However, this score is still more than 9 points lower than the model's performance on the regular evaluation.
Next, the 2-hop bi-attention model with the control unit obtains a higher EM score than the baseline in the adversarial evaluation, demonstrating better robustness against the adversaries. 
After this 2-hop model is additionally supervised to predict the sentence-level supporting facts, the performance in both regular and adversarial evaluation decreases a bit, but still outperforms both baselines in the regular evaluation (with stat. significance).
One possible explanation for this performance drop is that the 2-hop model without the extra task of predicting supporting facts overfits to the task of the final answer prediction, thus achieving higher scores.

\begin{table}[t]
\centering
\begin{small}
\begin{tabular}[t]{lcccc}
\toprule
\multirow{2}*{}& A4D-R & A4D-P & A8D-R & A8D-P \\
\midrule
1-hop Base & 37.65 & 37.72 & 34.14 & 34.84 \\
1-hop Base + sp& 44.65 & 44.51 & 43.42 & 43.59 \\
2-hop & 40.72 & 41.03 & 37.26 & 37.70 \\
2-hop + sp & \textbf{46.87} & \textbf{47.14} & \textbf{44.28} & \textbf{44.44}\\
\bottomrule
\end{tabular}
\vspace{-1pt}
\caption{EM scores on 4 adversarial evaluation settings after training on \textsc{Add4Docs-Rand}. `-R' and `-P' represent random insertion and prepending.
A4D and A8D stands for \textsc{Add4Docs} and \textsc{Add8Docs} adv-dev sets.
\label{tab:4adv-test}}
\vspace{-1pt}
\end{small}
\end{table}

\paragraph{Adversarially-Trained Models}
We further train all four models with the adversarial training set, and the results are shown in the last two columns in \tabref{tab:main}.
Comparing the numbers horizontally, we observe that after adversarial training, both the baselines and the 2-hop models with control unit gained statistically significant\footnote{Statistical significance of p $< 0.01$.} improvement on the adversarial evaluations.
Comparing the numbers in \tabref{tab:main} vertically, we show that the 2-hop model (row 3) achieves significantly (p-value $< 0.001$) better results than the baseline (row 1) on both regular and adversarial evaluation.
After we add the sentence-level supporting-fact supervision, the 2-hop model (row 4) obtains further improvements in both regular and adversarial evaluation.
Overall, we hope that these initial improvements will motivate the development of new models that combine explicit compositional reasoning with adversarial training.

\begin{table}[t]
\centering
\begin{small}
\begin{tabular}[t]{lcccc}
\toprule
Train & Regular & Regular & Adv & Adv\\
Eval & Regular & Adv & Regular & Adv \\
\midrule
    2-hop & \textbf{47.68} & \textbf{34.71} & \textbf{45.71} & \textbf{40.72} \\
    2-hop - Ctrl & 46.12 & 32.46 & 45.20 & 40.32 \\
    2-hop - Bridge & 43.31 & 31.80 & 41.90 & 37.37\\
    1-hop Base & 42.32 & 26.67 & 41.55 & 37.65 \\
\bottomrule
\end{tabular}
\vspace{-1pt}
\caption{Ablation for the Control unit and Bridge-entity supervision, reported as EM scores after training on the regular or adversarial \textsc{Add4Docs-Rand} data, and evaluation on regular dev set and \textsc{Add4Docs-Rand} adv-dev set. Note that 1-hop Base is same as 2-hop without both control unit and bridge-entity supervision.
    \label{tab:ablation}}
    \vspace{-1pt}
\end{small}
\end{table}

\paragraph{Adversary Ablation}
In order to test the robustness of the adversarially-trained models against new adversaries, we additionally evaluate them on dev sets with varying numbers of adversarial documents and a different adversary placement strategy elaborated in \secref{ssec:exp-adv-eval-train}. 
As shown in the first two columns in \tabref{tab:4adv-test}, neither the baselines nor the 2-hop models are affected when the adversarial documents are pre-pended to the context.
When the number of adversarial documents per supporting document with answer is increased to eight, all four models' performance drops by more than 1 points, but again the 2-hop model, with or without supporting-fact supervision, continues to outperform its single-hop counterpart.

\paragraph{Control Unit Ablation}
We also conduct an ablation study on the 2-hop model by removing the control unit. As shown in the first two rows of \tabref{tab:ablation}, the model with the control unit outperforms the alternative in all 4 settings with different training and evaluation data combinations. 
The results validate our intuition that the control unit can improve the model's multi-hop reasoning ability and robustness against adversarial documents.

\paragraph{Bridge-Entity Supervision Ablation}
We further investigate how intermediate supervision of finding the bridge entity affects the overall performance.
For this ablation, we also construct another 2-hop model without the bridge-entity supervision, using 2 unshared layers of bi-attention (2-hop - Bridge), as opposed to our previous model with 2 parallel, shared layers of bi-attention.
As shown in \tabref{tab:ablation}, both the 2-hop and 1-hop models without the bridge-entity supervision suffer large drops in the EM scores, suggesting that intermediate supervision is important for the model to learn the compositional reasoning behavior.

%% file: tex/analysis.tex
\section{Analysis}
In this section, we seek to understand the behavior of the model under the influence of the adversarial examples.
Following \citet{jia2017adversarial}, we focus on examples where the model predicted the correct answer on the regular dev set.
This portion of the examples is divided into ``model-successes" --- where the model continues to predict the correct answer given the adversarial documents, and ``model-failures" --- where the model makes the wrong prediction on the adversarial example.

\paragraph{Manual Verification of Adversaries}
\label{ssec:manual_veri}
We first verify that the adversarial documents do not contradict the original answer.
As elaborated in \secref{ssec:adv-const}, we assume that the bridge entity is the title of a supporting document and substitute it with another title sampled from the training/dev set.
Thus, the contradiction could arise when the adversarial document $p'_2$ is linked with $p_1$ with another entity other than the titles.
We randomly sample 50 examples in \textsc{Add4Docs-Rand}, and find 0 example where the fake answers in the adversarial docs contradict the original answer.
This shows that our adversary construction is effective in breaking the logical connection between the supporting documents and adversaries.

\paragraph{Model Error Analysis}
Next, we try to understand the model's false prediction in the ``model-failures" subset on \textsc{Add4Docs-Rand}.
For the 1-hop Baseline trained on regular data (2nd row, 2nd column in \tabref{tab:main}), in 96.3\% of the failures, the model's prediction spans at least one of the adversarial documents.
For the same baseline trained with adversarial data, the model's prediction spans at least one adversarial document in 95.4\% of the failures.
We further found that in some examples, the span predicted on the adversarial data is much longer than the span predicted on the original dev set, sometimes starting from a word in one document and ending several documents later. 
This is because our models predict the start and end indexes separately, and thus could be affected by different adversarial documents in the context.

\paragraph{Adversary Failure Analysis}
Finally, we investigate those ``model-successes", where the adversarial examples fail to fool the model.
Specifically, we find that some questions can be answered with a single document.
For the question ``Who produced the film that was Jennifer Kent's directorial debut?", one supporting document states ``The Babadook is ... directed by Jennifer Kent in her directorial debut, and produced by Kristina Tarbell and Kristian Corneille."
In this situation, even an adversary is unable to change the single-hop nature of the question.
We refer to the appendix for the full example. 

\paragraph{Toward Better Multi-Hop QA Datasets}
Lastly, we provide some intuition that is of importance for future attempts in collecting multi-hop questions. 
In general, the final sub-question of a multi-hop question should not be over-specific, so as to avoid large semantic match between the question and the surrounding context of the answer. 
Compared to the question in \figref{fig:example-intro}, it is harder to find a shortcut for the question ``What government position was held by the woman who portrayed Corliss Archer in ..." because the final sub-question (``What government position") contains less information for the model to directly exploit, and it is more possible that a distracting document breaks the reasoning shortcut by mentioning another government position held by a person.

%% file: tex/related.tex
\section{Related Works}
\paragraph{Multi-hop Reading Comprehension}
The last few years have witnessed significant progress on large-scale QA datasets including cloze-style blank-filling tasks~\citep{Hermann:cnndm}, open-domain QA~\citep{yang2015wikiqa}, QA with answer span prediction~\citep{rajpurkar2016squad,rajpurkar2018squad2}, and generative QA~\citep{nguyen2016msmarco}.
However, all of the above datasets are confined to a single-document context per question domain. 

Earlier attempts in multi-hop QA focused on reasoning about the relations in a knowledge base~\citep{Jain2016Question,zhou2018multirelation,lin2018multi} or tables \citep{yin2015neural}.
The bAbI dataset~\cite{weston2015babi} uses synthetic contextx and requires the model to combine multiple pieces of evidence in the text-based context. 
TriviaQA~\cite{joshi2017triviaqa} includes a small portion of questions that require cross-sentence inference.
\citet{welbl2017qangaroo} uses Wikipedia articles as the context and subject-relation pairs as the query, and construct the multi-hop QAngaroo dataset by traversing a directed bipartite graph. 
It is designed in a way such that the evidence required to answer a query could be spread across multiple documents that are not directly related to the query.
HotpotQA~\cite{yang2018hotpotqa} is a more recent multi-hop dataset that has crowd-sourced questions with diverse syntactic and semantic features. 
HotpotQA and QAngaroo also differ in their types of multi-hop reasoning covered.
Because of the knowledge-base domain and the triplet format used in the construction, QAngaroo's questions usually require inferring the desired property of a query subject by finding a bridge entity that connects the query to the answer.
HotpotQA includes three more types of question, each requiring a different reasoning paradigm. 
Some examples require inferring the bridge entity from the question (Type I in \citet{yang2018hotpotqa}), while others demand checking facts or comparing subjects' properties from two different documents (Type II and comparison question).

\paragraph{Adversarial Evaluation and Training}
\citet{jia2017adversarial} first applied adversarial evaluation to QA models on the SQuAD~\cite{rajpurkar2016squad} dataset by generating a sentence that only resembles the question syntactically and appending it to the paragraph. 
They report that the performances of state-of-the-art QA models~\cite{seo2016bidaf,hu2017reinforced,huang2017fusionnet} drop significantly when evaluated on the adversarial data.
\citet{wang2018robust} further improves the AddSent adversary and proposed AddSentDiverse that employs a diverse vocabulary for the question conversion procedure. 
They show that models trained with such adversarial examples can be robust against a wide range of adversarial evaluation samples. 
Our paper shares the spirit with these two works as we also try to investigate models' over-stability to semantics-altering perturbations.
However, our study also differs from the previous works~\cite{jia2017adversarial,wang2018robust} in two points.
First, we generate adversarial documents by replacing the answer and bridge entities in the supporting documents instead of converting the question into a statement.
Second, our adversarial documents still preserve words with common semantic meaning to the question so that it can distract models that are exploiting the reasoning shortcut in the context.

%% file: tex/conclusion.tex
\section{Conclusion}
In this work, we identified reasoning shortcuts in the HotpotQA dataset where the model can locate the answer without multi-hop reasoning. 
We constructed adversarial documents that can fool the models exploiting the shortcut, and found that the performance of a state-of-the-art model dropped significantly under our adversarial examples.
We showed that this baseline can improve on the adversarial evaluation after being trained on the adversarial data.
We next proposed to use a control unit that dynamically attends to the question to guide the bi-attention in multi-hop reasoning.
Trained on the regular data, this 2-hop model is more robust against the adversary than the baseline; and after being trained with adversarial data, this model achieved further improvements on the adversarial evaluation and also outperforms the baseline. Overall, we hope that these insights and initial improvements will motivate the development of new models that combine explicit compositional reasoning with adversarial training.

%% file: tex/acknowledgement.tex
\section{Acknowledgement}
We thank the reviewers for their helpful comments. This work was supported by DARPA (YFA17-D17AP00022), Google Faculty Research
Award, Bloomberg Data Science Research Grant, Salesforce Deep Learning Research Grant, Nvidia GPU awards, Amazon AWS, and Google Cloud Credits. The views, opinions, and/or findings contained in this article are those of the authors and should not be interpreted as representing the official views or policies, either expressed or implied, of the funding agency.

%% file: tex/appendix.tex
\appendix

\section*{Appendix}

\section{Examples}

\begin{figure}[ht]
\centering
\includegraphics[width=0.47\textwidth]{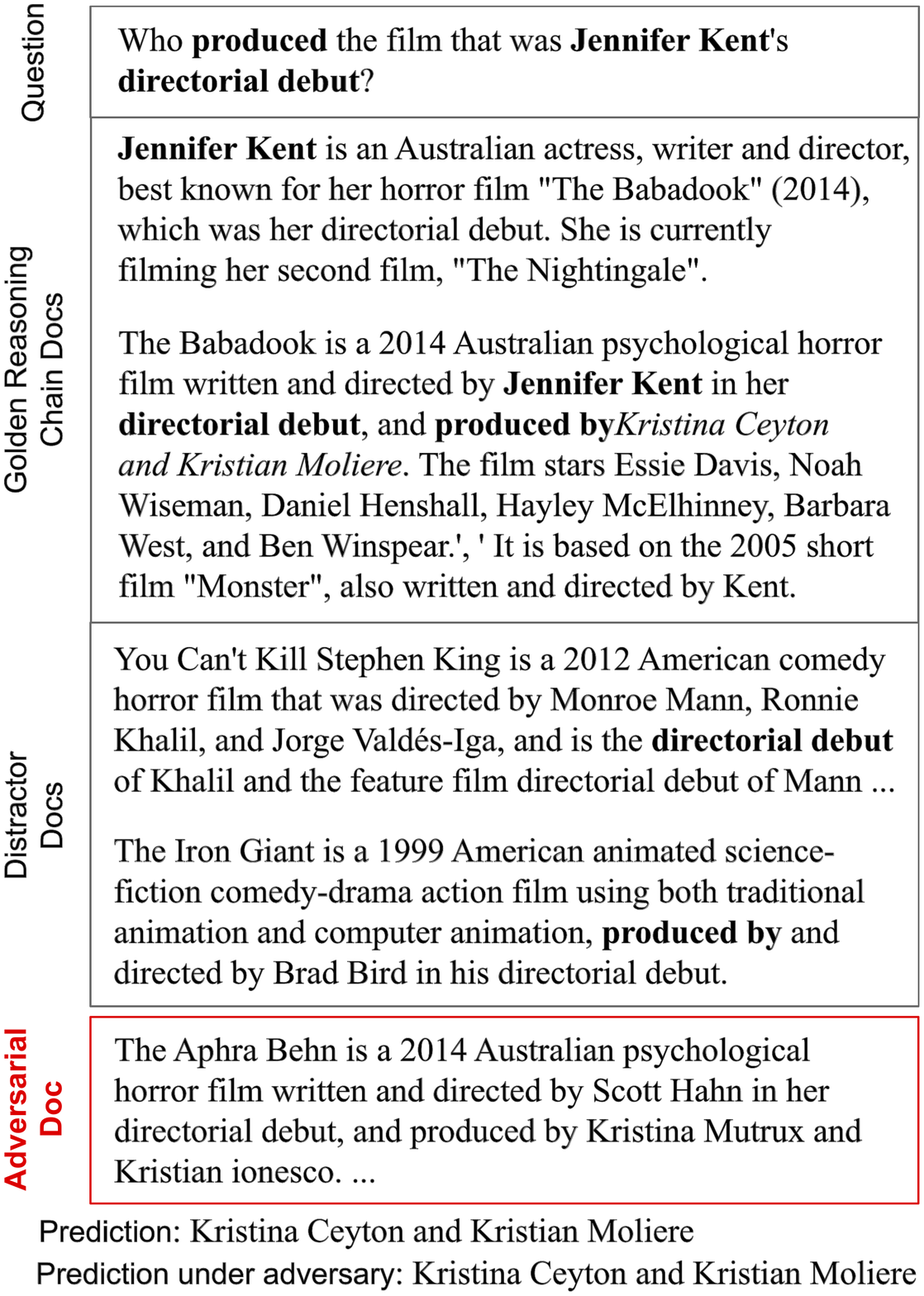}
\vspace{-1pt} 
\caption{A single-hop HotpotQA example that cannot be fixed with our adversary. 
 \vspace{-2pt}
\label{fig:example-adv-fail}}
\end{figure}

We show an HotpotQA~\cite{yang2018hotpotqa} example of our adversarial documents fail to fool the model into predicting the fake answer. As shown in \figref{fig:example-adv-fail}, the question can be directly answered by the second document in the Golden Reasoning Chain. Therefore, it is logically impossible to create an adversarial document to break this single-hop situation without introducing contradiction.

%% file: main.bbl
\begin{thebibliography}{28}
\expandafter\ifx\csname natexlab\endcsname\relax\def\natexlab#1{#1}\fi

\bibitem[{Chen et~al.(2017)Chen, Fisch, Weston, and Bordes}]{chen2017reading}
Danqi Chen, Adam Fisch, Jason Weston, and Antoine Bordes. 2017.
\newblock Reading wikipedia to answer open-domain questions.
\newblock In \emph{Proceedings of the 55th Annual Meeting of the Association
  for Computational Linguistics (Volume 1: Long Papers)}.

\bibitem[{Clark and Gardner(2018)}]{clark2017simple}
Christopher Clark and Matt Gardner. 2018.
\newblock Simple and effective multi-paragraph reading comprehension.
\newblock In \emph{Proceedings of the 56th Annual Meeting of the Association
  for Computational Linguistics (Volume 1: Long Papers)}.

\bibitem[{Efron and Tibshirani(1994)}]{Efron:94}
Bradley Efron and Robert~J Tibshirani. 1994.
\newblock \emph{An introduction to the bootstrap}.
\newblock CRC press.

\bibitem[{Hermann et~al.(2015)Hermann, Kocisky, Grefenstette, Espeholt, Kay,
  Suleyman, and Blunsom}]{Hermann:cnndm}
Karl~Moritz Hermann, Tomas Kocisky, Edward Grefenstette, Lasse Espeholt, Will
  Kay, Mustafa Suleyman, and Phil Blunsom. 2015.
\newblock Teaching machines to read and comprehend.
\newblock In \emph{Advances in Neural Information Processing Systems}, pages
  1693--1701.

\bibitem[{Hochreiter and Schmidhuber(1997)}]{hochreiter1997lstm}
Sepp Hochreiter and J{\"u}rgen Schmidhuber. 1997.
\newblock Long short-term memory.
\newblock \emph{Neural computation}, 9(8):1735--1780.

\bibitem[{Hu et~al.(2018)Hu, Peng, Huang, Qiu, Wei, and
  Zhou}]{hu2017reinforced}
Minghao Hu, Yuxing Peng, Zhen Huang, Xipeng Qiu, Furu Wei, and Ming Zhou. 2018.
\newblock Reinforced mnemonic reader for machine reading comprehension.
\newblock In \emph{IJCAI}.

\bibitem[{Huang et~al.(2018)Huang, Zhu, Shen, and Chen}]{huang2017fusionnet}
Hsin-Yuan Huang, Chenguang Zhu, Yelong Shen, and Weizhu Chen. 2018.
\newblock Fusionnet: Fusing via fully-aware attention with application to
  machine comprehension.
\newblock In \emph{ICLR}.

\bibitem[{Hudson and Manning(2018)}]{hudson2018compositional}
Drew~A Hudson and Christopher~D Manning. 2018.
\newblock Compositional attention networks for machine reasoning.
\newblock In \emph{Proceedings of ICLR}.

\bibitem[{Jain(2016)}]{Jain2016Question}
Sarthak Jain. 2016.
\newblock Question answering over knowledge base using factual memory networks.
\newblock In \emph{Proceedings of the NAACL Student Research Workshop}.
  Association for Computational Linguistics.

\bibitem[{Jia and Liang(2017)}]{jia2017adversarial}
Robin Jia and Percy Liang. 2017.
\newblock Adversarial examples for evaluating reading comprehension systems.
\newblock In \emph{Proceedings of the Conference on Empirical Methods in
  Natural Language Processing (EMNLP)}.

\bibitem[{Joshi et~al.(2017)Joshi, Choi, Weld, and
  Zettlemoyer}]{joshi2017triviaqa}
Mandar Joshi, Eunsol Choi, Daniel~S Weld, and Luke Zettlemoyer. 2017.
\newblock Triviaqa: A large scale distantly supervised challenge dataset for
  reading comprehension.
\newblock \emph{arXiv preprint arXiv:1705.03551}.

\bibitem[{Kingma and Ba(2014)}]{Kingma2014AdamAM}
Diederik~P. Kingma and Jimmy Ba. 2014.
\newblock Adam: A method for stochastic optimization.
\newblock \emph{CoRR}.

\bibitem[{Lin et~al.(2018)Lin, Socher, and Xiong}]{lin2018multi}
Xi~Victoria Lin, Richard Socher, and Caiming Xiong. 2018.
\newblock Multi-hop knowledge graph reasoning with reward shaping.
\newblock In \emph{EMNLP}.

\bibitem[{Nguyen et~al.(2016)Nguyen, Rosenberg, Song, Gao, Tiwary, Majumder,
  and Deng}]{nguyen2016msmarco}
Tri Nguyen, Mir Rosenberg, Xia Song, Jianfeng Gao, Saurabh Tiwary, Rangan
  Majumder, and Li~Deng. 2016.
\newblock Ms marco: A human generated machine reading comprehension dataset.
\newblock \emph{arXiv preprint arXiv:1611.09268}.

\bibitem[{Pennington et~al.(2014)Pennington, Socher, and
  Manning}]{pennington2014glove}
Jeffrey Pennington, Richard Socher, and Christopher~D Manning. 2014.
\newblock Glove: Global vectors for word representation.
\newblock In \emph{Conference on Empirical Methods in Natural Language
  Processing (EMNLP)}.

\bibitem[{Rajpurkar et~al.(2018)Rajpurkar, Jia, and
  Liang}]{rajpurkar2018squad2}
P.~Rajpurkar, R.~Jia, and P.~Liang. 2018.
\newblock Know what you don't know: Unanswerable questions for {SQuAD}.
\newblock In \emph{Association for Computational Linguistics (ACL)}.

\bibitem[{Rajpurkar et~al.(2016)Rajpurkar, Zhang, Lopyrev, and
  Liang}]{rajpurkar2016squad}
Pranav Rajpurkar, Jian Zhang, Konstantin Lopyrev, and Percy Liang. 2016.
\newblock Squad: 100,000+ questions for machine comprehension of text.
\newblock In \emph{Proceedings of the Conference on Empirical Methods in
  Natural Language Processing (EMNLP)}.

\bibitem[{Seo et~al.(2017)Seo, Kembhavi, Farhadi, and
  Hajishirzi}]{seo2016bidaf}
Minjoon Seo, Aniruddha Kembhavi, Ali Farhadi, and Hannaneh Hajishirzi. 2017.
\newblock Bidirectional attention flow for machine comprehension.
\newblock In \emph{International Conference on Learning Representations
  (ICLR)}.

\bibitem[{Srivastava et~al.(2015)Srivastava, Greff, and
  Schmidhuber}]{srivastava2015highway}
Rupesh~Kumar Srivastava, Klaus Greff, and J{\"{u}}rgen Schmidhuber. 2015.
\newblock Highway networks.
\newblock In \emph{International Conference on Machine Learning (ICML)}.

\bibitem[{Wang et~al.(2017)Wang, Yang, Wei, Chang, and Zhou}]{wang2017gated}
Wenhui Wang, Nan Yang, Furu Wei, Baobao Chang, and Ming Zhou. 2017.
\newblock Gated self-matching networks for reading comprehension and question
  answering.
\newblock In \emph{Proceedings of the 55th Annual Meeting of the Association
  for Computational Linguistics (Volume 1: Long Papers)}, volume~1, pages
  189--198.

\bibitem[{Wang and Bansal(2018)}]{wang2018robust}
Yicheng Wang and Mohit Bansal. 2018.
\newblock Robust machine comprehension models via adversarial training.
\newblock In \emph{NAACL}.

\bibitem[{Welbl et~al.(2017)Welbl, Stenetorp, and Riedel}]{welbl2017qangaroo}
Johannes Welbl, Pontus Stenetorp, and Sebastian Riedel. 2017.
\newblock Constructing datasets for multi-hop reading comprehension across
  documents.
\newblock In \emph{TACL}.

\bibitem[{Weston et~al.(2016)Weston, Bordes, Chopra, Rush, van Merri{\"e}nboer,
  Joulin, and Mikolov}]{weston2015babi}
Jason Weston, Antoine Bordes, Sumit Chopra, Alexander~M Rush, Bart van
  Merri{\"e}nboer, Armand Joulin, and Tomas Mikolov. 2016.
\newblock Towards ai-complete question answering: A set of prerequisite toy
  tasks.
\newblock In \emph{ICLR}.

\bibitem[{Xiong et~al.(2017)Xiong, Zhong, and Socher}]{xiong2016dynamic}
Caiming Xiong, Victor Zhong, and Richard Socher. 2017.
\newblock Dynamic coattention networks for question answering.
\newblock In \emph{ICLR}.

\bibitem[{Yang et~al.(2015)Yang, Yih, and Meek}]{yang2015wikiqa}
Yi~Yang, Wen-tau Yih, and Christopher Meek. 2015.
\newblock Wikiqa: A challenge dataset for open-domain question answering.
\newblock In \emph{Proceedings of the 2015 Conference on Empirical Methods in
  Natural Language Processing}, pages 2013--2018.

\bibitem[{Yang et~al.(2018)Yang, Qi, Zhang, Bengio, Cohen, Salakhutdinov, and
  Manning}]{yang2018hotpotqa}
Zhilin Yang, Peng Qi, Saizheng Zhang, Yoshua Bengio, William~W Cohen, Ruslan
  Salakhutdinov, and Christopher~D Manning. 2018.
\newblock Hotpotqa: A dataset for diverse, explainable multi-hop question
  answering.
\newblock In \emph{Conference on Empirical Methods in Natural Language
  Processing (EMNLP)}.

\bibitem[{Yin et~al.(2015)Yin, Lu, Li, and Kao}]{yin2015neural}
Pengcheng Yin, Zhengdong Lu, Hang Li, and Ben Kao. 2015.
\newblock Neural enquirer: Learning to query tables.
\newblock \emph{arXiv preprint}.

\bibitem[{Zhou et~al.(2018)Zhou, Huang, and Zhu}]{zhou2018multirelation}
Mantong Zhou, Minlie Huang, and Xiaoyan Zhu. 2018.
\newblock An interpretable reasoning network for multi-relation question
  answering.
\newblock In \emph{Proceedings of the 27th International Conference on
  Computational Linguistics}.

\end{thebibliography}
